\documentclass[10pt,twocolumn,letterpaper]{article}

\usepackage{cvpr}              % To produce the CAMERA-READY version
\usepackage[dvipsnames]{xcolor}

\definecolor{cvprblue}{rgb}{0.21,0.49,0.74}
\usepackage[pagebackref,breaklinks,colorlinks,citecolor=cvprblue]{hyperref}

\def\paperID{19} % *** Enter the Paper ID here
\def\confName{CVPR}
\def\confYear{2024}

\title{Zero-Shot Vision-and-Language Navigation with Collision Mitigation in Continuous Environment
}
\author{Seongjun Jeong$^{1}$\quad Gi-Cheon Kang$^{1,3}$\quad Joochan Kim  $^{2}$\quad Byoung-Tak Zhang $^{1,2,3}$\thanks{Corresponding author}\\
{\normalsize $^1$Interdisciplinary Program in Artificial Intelligence, Seoul National University}\\
{\normalsize $^2$Dept. of Computer Science and Engineering, Seoul National University}\\
{\normalsize $^3$AI Institute of Seoul National University (AIIS)} \\
{\tt\small {\{jsj4968,chonkang,tikatoka,btzhang\}@snu.ac.kr}}}
\begin{document}
\maketitle

\section{Introduction}

We explore Zero-Shot Vision-and-Language Navigation in Continuous Environment, where agents navigate using natural language instructions without any training data. Collecting instruction-path annotation data is an expensive task. Additionally, humans can navigate without prior learning about the environment. Equipping an embodied agent with this ability is an important task for creating a general-purpose agent that can perform tasks in a variety of unfamiliar environments. In discrete environments, Vision-and-Language Navigation (VLN)\cite{anderson2018vision} is performed through graph traversal, assuming collision-free movement between nodes. However, in continuous environments\cite{krantz2020beyond}, navigation must be done through low-level actions to the destination, considering possible collisions. 

We propose the zero-shot Vision-and-Language Navigation with Collision Mitigation (VLN-CM), which takes these considerations. VLN-CM is composed of four modules and predicts the direction and distance of the next movement at each step. We utilize large foundation models for each modules. To select the direction, we use the Attention Spot Predictor (ASP), View Selector (VS), and Progress Monitor (PM). The ASP employs a Large Language Model (e.g. ChatGPT\cite{ouyang2022training}) to split navigation instructions into attention spots, which are objects or scenes at the location to move to (e.g. a yellow door). The VS selects from panorama images provided at 30-degree intervals the one that includes the attention spot, using CLIP\cite{radford2021learning} similarity. We then choose the angle of the selected image as the direction to move in. The PM uses a rule-based approach to decide which attention spot to focus on next, among multiple spots derived from the instructions. If the similarity between the current attention spot and the visual observations decreases consecutively at each step, the PM determines that the agent has passed the current spot and moves on to the next one. For selecting the distance to move, we employed the Open Map Predictor (OMP). The OMP uses panorama depth information to predict an occupancy mask. We then selected a collision-free distance in the predicted direction based on the occupancy mask.

% We use the Attention Spot Predictor (ASP), View Selector (VS), and Progress Monitor (PM) to determine movement direction. The ASP uses a Large Language Model, like ChatGPT\cite{ouyang2022training}, to identify key objects or scenes, called attention spots, from navigation instructions. The VS then picks the best panoramic image, taken at 30-degree intervals, that shows the attention spot using CLIP\cite{radford2021learning} similarity, and we move in the direction of that image. The PM decides the next attention spot based on decreasing similarity at each step, indicating the agent has passed the current spot. For movement distance, the Open Map Predictor (OMP) assesses panorama depth to predict safe travel distances using an occupancy mask.

We evaluated our method using the validation data of VLN-CE\cite{krantz2020beyond}. Our approach showed better performance than several baseline methods, and the OPM was effective in mitigating collisions for the agent.

\section{Method}
\label{sec:method}

\begin{figure*}
    \centering
    \includegraphics[width=15cm]{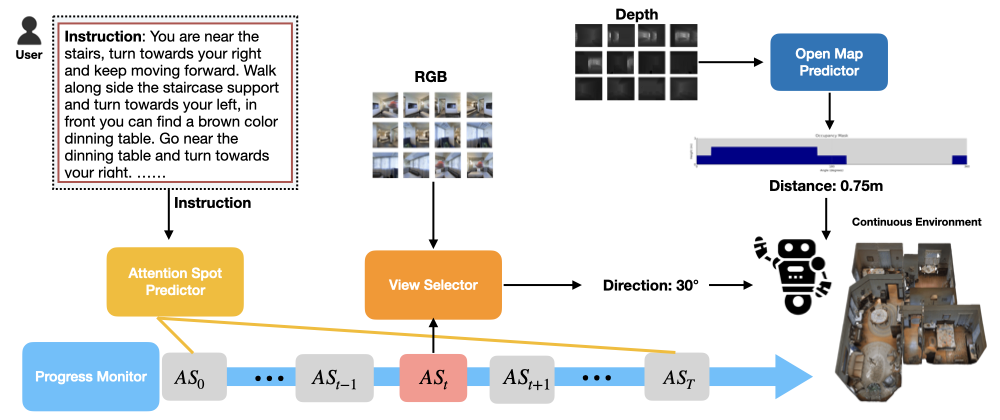}
    \caption{Overview of the VLN-CM}
    \label{fig:enter-label}
\end{figure*}

\subsection{Attention Spot Predictor}

The Attention Spot Predictor(ASP) decomposes the natural language instructions into specific attention spots, which are key visual markers within the environment, such as identifiable objects or unique scenes (e.g., a yellow door, a red chair). By parsing these complex instructions into simpler, actionable items, the ASP helps to guide the agent more effectively toward its goal. This module utilizes a Large Language Model (LLM), such as ChatGPT 3.5, for parsing complex navigation instructions into attention spots.

\subsection{View Selector}
The View Selector(VS) interacts directly with the egocentric views available to the agent, which are provided at regular 30-degree intervals. The VS employs the CLIP\cite{radford2021learning} to match these views with the attention spots identified by the ASP. By doing so, it selects the view that best corresponds to the next target location, effectively determining the direction in which the agent should head.

\subsection{Progress Monitor}

The Progress Monitor(PM) is a rule-based system that tracks the agent's progress towards each attention spot. It evaluates whether the agent is approaching or moving away from the attention spot by monitoring changes in the visual similarity between the attention spot and the agent's current views. If the similarity decreases consistently, the PM infers that the agent has passed the attention spot and updates the target to the next in line.

\subsection{Open Map Predictor}

To ensure safe navigation, the Open Map Predictor(OMP) uses the depth information to create an occupancy mask, which identifies areas that are free from obstacles. This module then calculates a safe and collision-free trajectory for the agent by determining how far it can move in the chosen direction before encountering a potential obstacle. It leverages a dataset based on the Habitat simulator to anticipate collision-free distances. For any chosen point in an open environment, we first collect its depth panoramas, which consist of 12 individual images taken at 30-degree intervals. The depth panoramas are input into the OPM, which predicts an occupancy mask covering 120 angles and 12 distances. We use transformer-based architecture \cite{hong2022bridging} for OPM

\section{Experiments}
\label{sec:experiments}
\subsection{Data and Evaluation Metrics}
We evaluate VLN-CM on the VLN-CE\cite{krantz2020beyond} unseen dataset. We measured its performance with Success Rate (SR), Success weighted by Inverse Path Length (SPL), and Collision Rate to assess both destination success, path efficiency, and the frequency of collisions. 

\subsection{Baselines}
We compare our model against two baseline agents:

\textbf{Random Agent:} The agent chooses actions based on observed training data probabilities—68\% move forward, 15\% turn left, 15\% turn right, 2\% stop—serving as a baseline for random decisions in navigation tasks.

\textbf{Hand-Crafted Agent:} The agent uses a basic navigation strategy by choosing a random direction, moving forward 37 times—the average trajectory length in the dataset—then stopping.

\subsection{Results}

Our VLN-CM model significantly outperformed baseline agents, achieving a SR of 0.11, compared to 0.03 for both Random and Hand-Crafted Agents.

Removing the OMP from VLN-CM resulted in a sharp drop in SR to 0.01 and increased collisions to 3.07. The most severe impact was observed when both OMP and ASP were omitted, leading to navigation failures and a collision rate of 24.49.

\begin{table}
  \centering
  \begin{tabular}{lcc}
    \hline
    \textbf{Model} & \textbf{\# SR} & \textbf{\# SPL} \\ \hline
    Random Agent & 0.03 & 0.02 \\
    Hand-Crafted Agent & 0.03 & 0.02 \\
    VLN-CM(ours) & 0.11 & 0.02 \\ \hline
\end{tabular}
\caption{Comparision with baselines}
\label{tab:baseline_comparison}
\end{table}

\begin{table}
  \centering
\begin{tabular}{lccc}
\hline
\textbf{Model} & \textbf{\# SR} & \textbf{\# SPL} & \textbf{\# Collsion} \\ \hline
VLN-CM & 0.11 & 0.02 & 0.67 \\
-OMP & 0.01 & 0.01 & 3.07\\
-ASP & 0 & 0 & 3.10\\
-OPN \& ASP & 0 & 0 & 24.49\\ \hline
\end{tabular}
\caption{Ablation Study.}
\label{tab:ablation}
\end{table}

\section{Conclusion}
\label{sec:conclusion}
The VLN-CM model significantly improves navigation in continuous environments with natural language, outperforming baselines in success rate and reducing collisions, highlighting the benefits of advanced modules like ASP and OMP for autonomous systems.

\noindent\textbf{Acknowledgements} This work was partly supported by the IITP (RS-2021-II212068-AIHub/10\%, RS-2021-II211343-GSAI/15\%, 2022-0-00951-LBA/15\%, 2022-0-00953-PICA/20\%), NRF (RS-2024-00353991-SPARC/20\%, RS-2023-00274280/10\%), and KEIT (RS-2024-00423940/10\%) grant funded by the Korean government.

{
    \small
    \bibliographystyle{ieeenat_fullname}
    \bibliography{main}

\begin{thebibliography}{5}
\providecommand{\natexlab}[1]{#1}
\providecommand{\url}[1]{\texttt{#1}}
\expandafter\ifx\csname urlstyle\endcsname\relax
  \providecommand{\doi}[1]{doi: #1}\else
  \providecommand{\doi}{doi: \begingroup \urlstyle{rm}\Url}\fi

\bibitem[Anderson et~al.(2018)Anderson, Wu, Teney, Bruce, Johnson, S{\"u}nderhauf, Reid, Gould, and Van Den~Hengel]{anderson2018vision}
Peter Anderson, Qi Wu, Damien Teney, Jake Bruce, Mark Johnson, Niko S{\"u}nderhauf, Ian Reid, Stephen Gould, and Anton Van Den~Hengel.
\newblock Vision-and-language navigation: Interpreting visually-grounded navigation instructions in real environments.
\newblock In \emph{Proceedings of the IEEE conference on computer vision and pattern recognition}, pages 3674--3683, 2018.

\bibitem[Hong et~al.(2022)Hong, Wang, Wu, and Gould]{hong2022bridging}
Yicong Hong, Zun Wang, Qi Wu, and Stephen Gould.
\newblock Bridging the gap between learning in discrete and continuous environments for vision-and-language navigation.
\newblock In \emph{Proceedings of the IEEE/CVF Conference on Computer Vision and Pattern Recognition}, pages 15439--15449, 2022.

\bibitem[Krantz et~al.(2020)Krantz, Wijmans, Majumdar, Batra, and Lee]{krantz2020beyond}
Jacob Krantz, Erik Wijmans, Arjun Majumdar, Dhruv Batra, and Stefan Lee.
\newblock Beyond the nav-graph: Vision-and-language navigation in continuous environments.
\newblock In \emph{Computer Vision--ECCV 2020: 16th European Conference, Glasgow, UK, August 23--28, 2020, Proceedings, Part XXVIII 16}, pages 104--120. Springer, 2020.

\bibitem[Ouyang et~al.(2022)Ouyang, Wu, Jiang, Almeida, Wainwright, Mishkin, Zhang, Agarwal, Slama, Ray, et~al.]{ouyang2022training}
Long Ouyang, Jeffrey Wu, Xu Jiang, Diogo Almeida, Carroll Wainwright, Pamela Mishkin, Chong Zhang, Sandhini Agarwal, Katarina Slama, Alex Ray, et~al.
\newblock Training language models to follow instructions with human feedback.
\newblock \emph{Advances in neural information processing systems}, 35:\penalty0 27730--27744, 2022.

\bibitem[Radford et~al.(2021)Radford, Kim, Hallacy, Ramesh, Goh, Agarwal, Sastry, Askell, Mishkin, Clark, et~al.]{radford2021learning}
Alec Radford, Jong~Wook Kim, Chris Hallacy, Aditya Ramesh, Gabriel Goh, Sandhini Agarwal, Girish Sastry, Amanda Askell, Pamela Mishkin, Jack Clark, et~al.
\newblock Learning transferable visual models from natural language supervision.
\newblock In \emph{International conference on machine learning}, pages 8748--8763. PMLR, 2021.

\end{thebibliography}
}


\end{document}